\documentclass[runningheads]{llncs}
\usepackage[T1]{fontenc}
\usepackage{graphicx}
\usepackage{color}
\usepackage{tikz}
\usepackage{pgfplots}
\pgfplotsset{compat=1.18}\usepgfplotslibrary{statistics}

\usepackage{comment}
\usepackage{wrapfig}

\begin{document}
\title{Automated Archival Descriptions with \\Federated Intelligence of LLMs}
%
%
\author{Jinghua Groppe\inst{1}\orcidID{0009-0009-0295-7029} \and
Andreas Marquet\inst{2} \orcidID{0000-0001-7238-9033} \and
Annabel Walz\inst{2} \orcidID{0000-0001-6894-8568} \and 
Sven Groppe \inst{1}\orcidID{0000-0001-5196-1117}
}
\authorrunning{J. Groppe, A. Marquet, A. Walz and S. Groppe}
%
\institute{IFIS, University of Lübeck, Ratzeburger Allee 160, 23562 Lübeck,
Germany \\
\email{jinghua.groppe@uni-luebeck.de}, \email{sven.groppe@uni-luebeck.de}\\
\and
Friedrich-Ebert-Stiftung e.V., Godesberger Allee 149, 53175 Bonn, Germany\\
\email{Andreas.Marquet@fes.de}, \email{Annabel.Walz@fes.de}}
\maketitle              
\begin{abstract}

Enforcing archival standards demands specialized expertise, and manually creating metadata descriptions for archival materials requires considerable manual effort and is error-prone. This work aims to explore the potential of agentic AI and large language models (LLMs) in addressing the challenges of implementing a standardized archival description process. To this end, we introduce an agentic AI-driven system for the automated generation of high-quality metadata descriptions of archival materials. We develop a federated optimization approach that unites the intelligence of multiple LLMs to construct optimal archival metadata. We also suggest methods to overcome the challenges associated with using LLMs for consistent metadata generation. To evaluate the feasibility and effectiveness of our techniques, we conducted extensive experiments using a real-world dataset of archival materials, which covers a variety of document types and formats. The evaluation results demonstrate the feasibility of our techniques and highlight the superior performance of the federated optimization approach compared to single-model solutions in metadata quality and reliability.

\keywords{information extraction \and file repositories \and archive \and metadata generation \and federated \and LLM \and agentic AI}
\end{abstract}

\section{Introduction}

Despite the availability of numerous specialized applications and document management systems, file repositories remain widely used technical environments that often merit preservation. From an archival perspective, the main challenge stems from their inherently flexible use by both individuals and organizations. This flexibility frequently leads to unstructured storage practices, inconsistent file naming conventions, redundancies, and incomplete or missing metadata, and these factors complicate efforts to maintain order and accessibility of file repositories. Additionally, the absence of a standardized procedure for file creation often results in the loss of contextual information, making it difficult to determine the origins and significance of stored files~\cite{wendt2017,naumann2017kreative,jaillant2022born}. In addition, file repositories generally lack quantitative and qualitative constraints, meaning that relevant information can be stored in unrestricted and highly variable forms~\cite{gillner2023abfragen,barrueco2022digital}.

Given the vast and practically unlimited volume of data~\cite{Groppe2020BigData}, coupled with the fundamentally different nature of digital media, traditional archival methods - such as appraisal, arrangement, and description - are only partially applicable in their current form. As a result, methodological and technical adaptations are necessary to ensure effective archival practices. 
Initial pilot projects relied predominantly on manual and intellectual efforts to process file repositories, further confirming the assumption that the associated workload is overwhelming, given the sheer size of the records.
Considering these challenges and the immense number of files to be managed, research topics such as the automatic extraction of metadata have become increasingly significant.

For efficient subject-specific processing, (semi-) automatic approaches - computational methods initiated, supervised, and refined by archivists - appear to be the most promising (e.g., \cite{Lenartz2022}). These (mostly) automated methods help manage the large-scale processing of files. However, intellectual input remains essential, particularly for contextualizing individual documents, understanding the workflows of the organizations or individuals creating them, and integrating files within existing archival holdings. These challenges are further exacerbated by the prevalence of hybrid record-keeping practices, in which analogue and digital documents coexist and their proportions shift over time. As a result, file systems present multiple complexities, introducing potential ambiguities regarding their origin, acquisition, and subsequent processing.

The use of artificial intelligence (AI) in archival practice has recently been tested for a variety of purposes, demonstrating significant potential - particularly in recognizing objects, people, or buildings in photographs and in content indexing of audiovisual material. Research conducted within the British LUSTRE network highlights both the promising applications of AI and the organizational prerequisites tied to adopting this emerging technology~\cite{Jaillant2022}. In a related initiative, the British National Archives evaluated several AI-based tools 
to support the selection and acquisition of digital records~\cite{archives2021using}. Building on these developments, this paper, centered on the file storage scenario, seeks to offer a more flexible and adaptive AI framework that advances the current state of the art.

Large language models (LLMs), owning to their remarkable capacity to produce human-like, context-aware responses across a wide range of tasks such as translation, summarization, question-answering, creative writing, and code generation, are the basis of chatbots (for general purposes like ChatGPT, Grok, DeepSeek and Gemini, and for special purposes like~\cite{Kessel2025Chatbot}) and are becoming an integral part of many aspects of daily life~\cite{Khorashadizadeh2024LLMKG}. Their ability to perform tasks with little to no prior examples - known as zero-shot or few-shot learning \cite{NEURIPS2020_1457c0d6} - makes them highly adaptable for use in data pipelines~\cite{Junior2024LLMDataPipeline,Kessel2025Analysis}. This flexibility coupled with other strengths is particularly valuable in generating archival metadata across heterogeneous documents, where LLMs can automate much of the process while allowing archivists to intervene with minimal effort to refine results and streamline subsequent workflows.

In this work, we investigate the abilities of agentic AI and LLMs in addressing challenges in implementing archival standards and propose a universal LLM-agent-driven framework for automated generation of archival metadata. Our main contributions include:
\begin{itemize}
\item an agentic AI-based metadata generation system, which utilizes the federated intelligence of multiple LLMs to automatically generate high-quality metadata for archival materials,
\item a methodology for determining best LLMs for archival metadata extraction,
\item techniques to address the challenges of using LLMs in generating consistent metadata descriptions,  
\item a federated optimization approach, which synthesizes an optimal metadata description from the results of an ensemble of LLMs, and
\item an extensive experimental evaluation on real-world archival materials containing documents of various types and in various formats. The evaluation results confirm the practicality of the LLM-agent-driven framework and demonstrate the superior performance of the federated approach compared to the use of individual LLMs alone.
\end{itemize}

\section{Related Work}

Large language models have already been used very successfully for information extraction tasks in other application areas: For example, Wang et al. \cite{Wang2023Code4Struct} and Parekh et al. \cite{Parekh2023} used OpenAI's large language models to extract structured data about events from unstructured text. Goel et al. \cite{Goel23} combined large language models with human expertise to annotate patient-related information in medical texts. By using state-of-the-art large language models such as GPT-4, Schimmenti et al. \cite{Schimmenti24} achieved very high accuracies for extracting metadata from historical texts (title: 98\%, type: 94\%, date: 89\%, location: 95\%, author: 79\%). Although the results of these studies come from partly different application areas, they indicate that large language models can also be successfully used for information extraction in archives, with their advantages in terms of language understanding and few-shot learning.

The transfer of document content into knowledge graphs can also be largely automated using large language models \cite{Khorashadizadeh2024Canocanilization,Ezzabady2024KGConstruct}, whereby archivists, for example, gain better control (compared to e.g. retrieval augmented generation) over what is freely accessible to archive users in a subsequent step, or which documents are subject to a protection period, by post-processing the knowledge graph.

The authors in \cite{ChenLLMEnsembles} discuss the related work about LLM ensembles. In this work, we propose a variant of LLM ensembles that uses an ensemble after inference approach to re-generate in case of unsuccessful validation and to generate a synthesized output from all outputs of the LLMs in the ensemble, tailored to the needs of archives by providing specialized contexts and a specialized validator.

None of the existing approaches investigates the use of large language models in the context of supporting archivists for archiving files.

\section{LLM-Agent-Driven Automatic Generation of Archival Metadata}\label{sec:Approach}
We suggest an agentic AI-based metadata generation system, which employs the federated intelligence of LLMs to automatically create a complete and precise metadata description of archival materials. We first introduce the agentic AI architecture of the LLM-based metadata generation system in Sec.~\ref{sec:archi}, then present the methodology to determine the best LLMs for the metadata generation task in Sec.~\ref{sec:candidate}. Sec.~\ref{consistent} describes the techniques to address the challenges of LLMs in constructing consistent metadata descriptions and Sec.~\ref{federated} presents the federated optimization approach, which leverages the intelligence of multiple LLMs to create high-quality archival metadata.

\subsection{System Architecture} 

\begin{figure}
    \centering
    \includegraphics[width=0.6\linewidth]{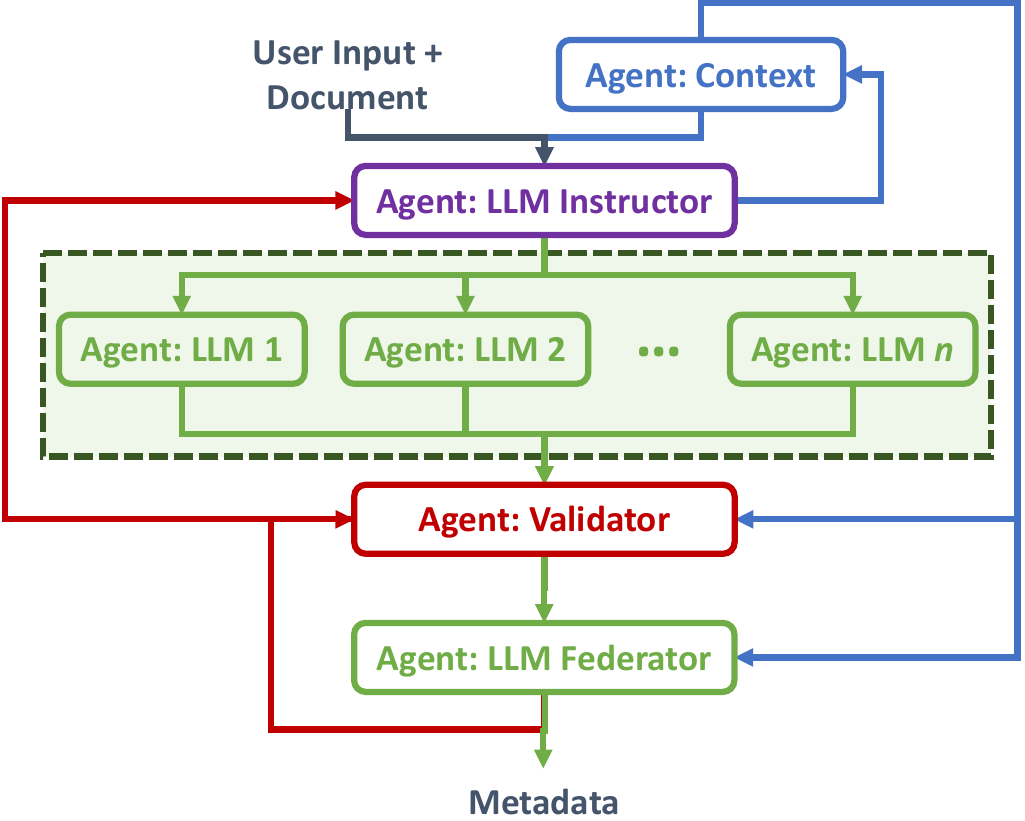}
    \caption{Agentic AI-driven system architecture for automated generation of optimal archival metadata}
    \label{fig:architecture}
\end{figure}


\label{sec:archi}
Fig.~\ref{fig:architecture} depicts the architecture of the automatic metadata generation system composed of agents powered by LLMs.  
After getting the user input, the LLM Instructor agent first analyses the input content and retrieves the corresponding context (like metadata specification of ISAD(G), Records in Contexts, and alike) from the Context agent (See Sec.~\ref{consistent} for details about contexts). Based on the user input and the context, the LLM Instructor then constructs instructions to generate metadata descriptions. These instructions and the archival material provided by the user are then fed into an ensemble of agents powered by various LLMs, and each LLM will generate a metadata description for the archival material according to the instructions. All the metadata descriptions generated are then checked by a Validator agent in terms of, e.g. the structure, completeness of extraction, and format adherence according to the context information. 

If validation fails, the validator agent will notify the LLM instructor agent of which LLMs output incorrect metadata descriptions. The LLM instructor then instructs the LLM agents to re-extract the metadata description. When all metadata descriptions created by the LLM agents in the ensemble are successfully validated, the validator agent sends all data and information to the LLM Federator agent. The Federator agent will construct synthesis instructions based on the requirements given in the context information and then synthesize an optimal metadata description from all individual ones according to the instructions. The new metadata description will be validated according to the context as well by the Validator agent. If validation fails, it will inform the LLM Federator agent to perform the synthesis again until validation is successful. 

Finally, the agentic AI system will output a high-quality metadata description for the given archival material in the required format, which maximizes the completeness and accuracy of the content and is optimally aligned with the archiving standard. 
The archival community widely adopts XML for storing metadata descriptions, as XML can present semi-structured data in a human-readable and machine-processable format while various optimization techniques \cite{groppe2008speeding,groppe2008filtering,groppe2006prototype,groppe2005schema,groppe2006xpath,groppe2006reformulating,groppe2006satisfiability,groppe2011transforming,groppe2003xpath} 
exist to speed up XML querying and transformations between different application-specific XML formats. 

\subsection{Determining Candidate LLMs} \label{sec:candidate}

A large number of LLMs are available for data extraction, and this presents an initial challenge: selecting the most suitable models to generate high-quality, reliable metadata descriptions. This section proposes a systematic methodology to identify the optimal candidate LLMs. 

Given a collection of LLMs: 
\begin{enumerate}
\item For each LLM, check whether it knows the standard to be used for the archival description. If this is not the case, it is removed from the collection. 
\item For each LLM, ask it to create a metadata description according to the archiving standard for some samples of the archival materials. If it is unable to do so or the quality of the extracted metadata description is low, it will be removed.
\item Design a prompt that exactly describes the format and structure of the metadata. For each LLM, we test with examples to see if it can extract metadata according to the requirements. If it cannot, it is removed. 
\item Evaluate the LLMs in the collection with samples of archival materials. If the performance of an LLM is lower than a threshold, it is removed.
\item Finally, if there are multiple LLMs from the same series in the collection, we will only keep the latest or more powerful one.
\end{enumerate}

The website \url{https://lmarena.ai/}\footnote{visited on 31.3.2025} collects 94 LLMs for researchers and practitioners to compare and test their performance, and so we use the collection as the initial one. After applying the selection methodology, we determine four LLMs: Grok 3 built by xAI, GPT-4-turbo by OpenAI, DeepSeek-V3 from DeepSeek, and Gemini 2.0 Flash from Google.

\subsection{Computing Consistent Metadata Descriptions} \label{consistent}
Many archival description standards provide descriptive elements, and most of them are made up of several words. This causes several problems: i) Some data store formats (like XML) require one-word tags. ii) Using long descriptive elements (or their concatenation) as tags is inefficient. For these reasons, simplified one-word tags are applied instead of descriptive names, and LLMs also follow this practice. However, different LLMs could adopt different ways to tag these elements, and even the same LLM might use various tags at different time. 
Furthermore, although we use the LLMs, which are familiar with the archival description standards, they do not always strictly follow them, and the metadata descriptions created by LLMs could have different structures and elements at different times. 
Metadata descriptions with varying structures and different tags are useless and need significantly post-processing. However, an automatic post-processing, which converts them into a consistent format, is almost impossible because the structure and terminology that LLMs use could be unlimited. Therefore, we need a solution to ensure that metadata descriptions generated by any LLMs at any time are strictly consistent in terms of structure and tags.

In this work, we suggest a context-based solution: We use a context that contains the necessary information to create consistent metadata. We provide the context to the LLMs, which use it to create metadata descriptions. We also provide the context to the validator agent, which will validate if the generated metadata descriptions comply with the specification given in the context. If validation fails, we ask the LLMs to re-regenerate the metadata description. This process is repeated until validation succeeds as shown in Fig.~\ref{fig:architecture}. In extreme cases, it could occur that LLMs cannot create a metadata description according to the requirements. But this can be easily handled by setting up a limited number of repetitions. If this number is exceeded, the system notifies the user.
 
To create a clear reference for LLMs to follow, such a context should include the following pieces of information:
\begin{itemize}
    \item the structure of the metadata description, which LLMs should follow.
    \item the purpose of each element, which provides LLMs with information on how to extract its value.
    \item examples for each element, which illustrate the use of the element.
    \item one-word tag for each element, which enables the use of uniform vocabulary.
\end{itemize}

\subsection{Creating Optimal Metadata Descriptions} \label{federated}
According to the methodology suggested in Sec~\ref{sec:candidate}, we can determine the best LLMs for automated generation of archival metadata. We want to utilize all of their intelligence to create an optimal archival metadata description. To realize this, we first ask each of the LLMs to analyze the archival material and generate metadata descriptions for it. Next, we aim to integrate the strengths of these metadata descriptions to synthesize a new, optimized version—one that maximizes completeness and precision of content while aligning closely with the metadata standard and the overarching intent of the archive.



To create the optimal metadata set from the source metadata ones, we suggest a systematic optimization strategy as follows:

\begin{enumerate}
    \item Analyze source metadata sets and identify their strengths and weaknesses.
    \begin{itemize}
    \item assess against the archival description standard: check if the metadata sets reflect the purposes of each element and follow the rules and conventions of the standard.
    \item compare against the archival material: cross-reference the metadata sets with the archival material to check if they reflect the archival intent and verify factual details like dates, creators, and content.
    \item identify variations, differences, and contradictions across the source metadata sets.
\end{itemize}
    
\item Element-by-element optimization

For each element:
\begin{itemize}
\item select the best base: Choose the most accurate or detailed entry from the source metadata sets as a starting point.
\item enhance with details: Add specifics from other sets or the archival material.
\item resolve discrepancies: cross-reference the metadata with the archival material and the archival standard.
\item correct errors: Adjust based on evidence (e.g., OCR errors given in notes).
\item standardize format: Ensure compliance with the rules of the archival standard.
\end{itemize}    
\item Validation and refinement
\begin{itemize}
 \item Cross-check with the archival standard and format requirements: Ensure that all elements follow the hierarchical structure and use the correct tags.
\item Align with archival material: Verify if the metadata reflects the intent and focus of the material.
\item Eliminate redundancy: Streamline overlapping details.
\item Enhance utility: Add details to aid users and systems if necessary.
\end{itemize}
\end{enumerate}

For automation, we need an agent to carry out these tasks. This agent should have knowledge of the archiving process, be familiar with the archival description standard, and have the ability to analyze and reason. No doubt, LLMs are currently an optimal candidate for such an agent. In order to find an LLM that can best perform the optimization task, we tested and evaluated a number of LLMs, and the results of the evaluation show that Grok3 from xAI is superior to other LLMs.


\section{Experimental Evaluation}

\subsection{Metrics and approach}

The quality of archival metadata is determined by many aspects: factual accuracy, content completeness, equivalence of intent, avoidance of contradiction, and alignment with the archival description standard. To evaluate these aspects, we need to compare the metadata sets against both the ground truth and the standard. The existing evaluation techniques, from the purely statistical scorers (Blue, ROUGE, Meteor and Levenshtein distance) to embedding models (like BertScore and MoverScore) to Natural Language Inference models (like NLI scorer) and BLEURT (which uses pre-trained models like BERT to score LLM outputs on ground truth), are not able to do this. 

Recently, LLM-as-a-Judge is emerging to perform complex evaluation due to their superior reasoning capabilities and knowledge about the world, and so we also use LLMs as evaluators. We instruct LLMs to evaluate the quality of the automatically generated metadata sets based on the ground truth and the standards, and compute a score for each element on a scale from 0 to 1 based on the factual accuracy, content completeness, contextual consistency, equivalence of intent, avoidance of contradiction and alignment with the archival description standard, where 0 indicates no similarity and 1 indicates identical or near-identical meaning, with increments reflecting nuanced differences. More concretely:

\begin{itemize}
\item 1: Exact or near-exact match in meaning and detail.
\item 0.75-0.99: High similarity with minor omissions or slight rephrasing.
\item 0.5-0.74: Moderate similarity with noticeable differences or partial matches.
\item 0.25-0.49: Low similarity with significant deviations or missing key details.
\item 0-0.24: Complete mismatch or entirely missing relevant information.
\end{itemize}

In order to find the most appropriate LLM as evaluator, we test and evaluate different LLMs, and the test results show that Grok3 is currently the best LLM for this task.

\subsection{Data and Ground Truth}

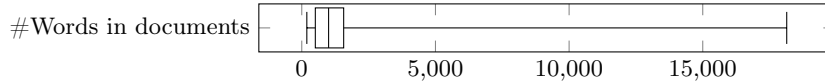
\begin{figure}  
\begin{tikzpicture}
\begin{axis}
[
    ytick={1},
    yticklabels={\#Words in documents},
    y post scale=0.1,
    width=\textwidth-3cm,
    clip=false,
    scaled y ticks=false,
    scaled x ticks=false
]
\addplot+[draw=black,solid,
    boxplot prepared={
        median=1003.5,
        upper quartile=1567.25,
        lower quartile=506.5,
        upper whisker=18141.0,
        lower whisker=189.0
    },
    ] coordinates {};
\end{axis}
\end{tikzpicture}
\caption{Box plots of number of words of each document in the evaluation dataset}\label{fig:nrOfWordsVersusScores}
\end{figure}

In our evaluation, we use real-world archive materials from Deutscher Gewerkschaftsbund (DGB)\footnote{The DGB is an umbrella organization for eight member unions in Germany. Please see \url{https://en.dgb.de/} (visited on 31.3.2025) for further information.} held at the Archive of Social Democracy of the Friedrich-Ebert-Stiftung (AdsD) (https://www.fes.de/archiv-der-sozialen-demokratie/). To reflect the diversity and heterogeneity of archival materials and evaluate the capabilities of LLMs in generating metadata descriptions, we choose 22 representative archival units from the archival corpus as evaluation data, which contain different document types (minutes, speeches, letters, memos, articles, e-mails, newsletters, presentations, notes, calendar, wage and income tax statistics) and cover various topics in different data formats (PDF, MS Word, PowerPoint, Excel).
Fig~\ref{fig:nrOfWordsVersusScores} presents the word statistics of the documents in the evaluation data set.  The number of words is on average 2037.25, the median 1003.5, the minimum 189, and the maximum 18141. Most documents have between 500 and 1500 words.

\begin{table}[!tb] 
\caption{ISAD(G) areas and elements considered in this evaluation with abbreviations used in this paper 
}\label{tab:ISADG}
\centering
\scriptsize
\begin{tabular}{| p{.245\textwidth} | p{.525\textwidth} | p{.21\textwidth} |}
\hline
\textbf{Area}/Element & \textbf{Description} & \textbf{Abbreviation} \\
\hline\hline
\textbf{1. Identity Statement Area} & This area provides essential information to identify the archival unit being described. & Id \\\hline
1.2 Title & The name given to the archival unit, either a formal title (if one exists) or a descriptive title created by the archivist
 & Id:Title \\
1.3 Date(s) & The date range or specific dates of the archival unit, including creation or accumulation dates. 
& Id:Date \\
1.4 Level of Description & Indicates the hierarchical level of the archival unit (e.g., fonds, sub-fonds, series, file, item). 
& Id:DescLev \\
1.5 Extent and medium of the unit of description  (quantity, bulk, or size) &  Describes the physical or digital size and format of the unit (e.g., "3 boxes," "0.5 linear meters," "10 digital files (PDF)"). 
& Id:Extent \\
\hline\hline
\textbf{3. Content and Structure Area} & This area describes the intellectual content and organization of the archival materials. & Cont \\\hline
3.1 Scope and Content & A summary of the subject matter, themes, and types of records included in the unit, helping users assess its relevance.
& Cont:Scope \\
\hline\hline
\textbf{4. Conditions of Access and Use Area} & This area outlines the terms and conditions for accessing and using the materials. & AccessUse \\\hline
4.3 Language(s) and Script(s) of Material & Identifies the language(s) and script(s) used in the materials (e.g., "English, with some documents in French; Latin script"). & AccessUse:Lang \\
4.4 Physical Characteristics and Technical Requirements & Describes any physical or technical conditions affecting use, such as fragility or digital file formats.
& AccessUse:PhysTech \\
\hline
\end{tabular}
\end{table}

\pgfplotstableread[row sep=\\,col sep=&]{
    Element            & LLM1 & LLM2 & LLM3 & LLM4 & Federated \\
Id:Title & 0.8931818181818181 & 0.7999999999999998 & 0.8863636363636362 & 0.8166666666666664 & 0.89 \\
Id:Date & 0.788095238095238 & 0.82 & 0.8547619047619047 & 0.8274999999999999 & 0.9052631578947368 \\
Id:DescLev & 0.9800000000000001 & 0.9785714285714286 & 0.8733333333333335 & 0.8933333333333333 & 0.9785714285714286 \\
Id:Extent & 0.8571428571428573 & 0.8074999999999999 & 0.9166666666666665 & 0.7952380952380954 & 0.836842105263158 \\
Cont:Scope & 0.891904761904762 & 0.7474999999999999 & 0.7952380952380954 & 0.773684210526316 & 0.8978947368421054 \\
AccessUse:Lang & 0.95 & 0.946875 & 0.9882352941176471 & 0.996875 & 0.9500000000000001 \\
AccessUse:PhysTech & 0.8441176470588235 & 0.8093750000000001 & 0.8705882352941179 & 0.615625 & 0.8733333333333335 \\
Id & 0.869431818181818 & 0.8387714285714284 & 0.8875000000000001 & 0.8241666666666667 & 0.8937499999999998 \\
Cont & 0.891904761904762 & 0.7474999999999999 & 0.7952380952380954 & 0.773684210526316 & 0.8978947368421054 \\
AccessUse & 0.9005882352941178 & 0.8812500000000001 & 0.9323529411764704 & 0.80625 & 0.9149999999999997 \\
All & 0.8860704545454545 & 0.8191166666666666 & 0.8659500000000001 & 0.8068619047619049 & 0.9017425000000001 \\
    }\mydata

\begin{figure}[h!]
\begin{tikzpicture}
    \begin{axis}[
            ybar=1.5pt,
            bar width=.1cm,
            width=\textwidth,
            height=.5\textwidth,
            legend style={at={(0.5,1.175)},
                anchor=north,legend columns=-1},
            symbolic x coords={Id:Title,Id:Date,Id:DescLev,Id:Extent,Cont:Scope,AccessUse:Lang,AccessUse:PhysTech,Id,Cont,AccessUse,All},
            xticklabel style={rotate=90},
            xtick=data,
            ylabel near ticks,
            ymin=0,ymax=1,
            ylabel={Score},
        ]
        \addplot table[x=Element,y=LLM1]{\mydata};
        \addplot table[x=Element,y=LLM2]{\mydata};
        \addplot table[x=Element,y=LLM3]{\mydata};
        \addplot table[x=Element,y=LLM4]{\mydata};
        \addplot table[x=Element,y=Federated]{\mydata};
        \legend{LLM 1, LLM 2, LLM 3, LLM 4, Federated}
    \end{axis}
\end{tikzpicture}
\caption{LLM scores for information extraction (average over all scores for all documents)}\label{fig:totaleval}
\end{figure}
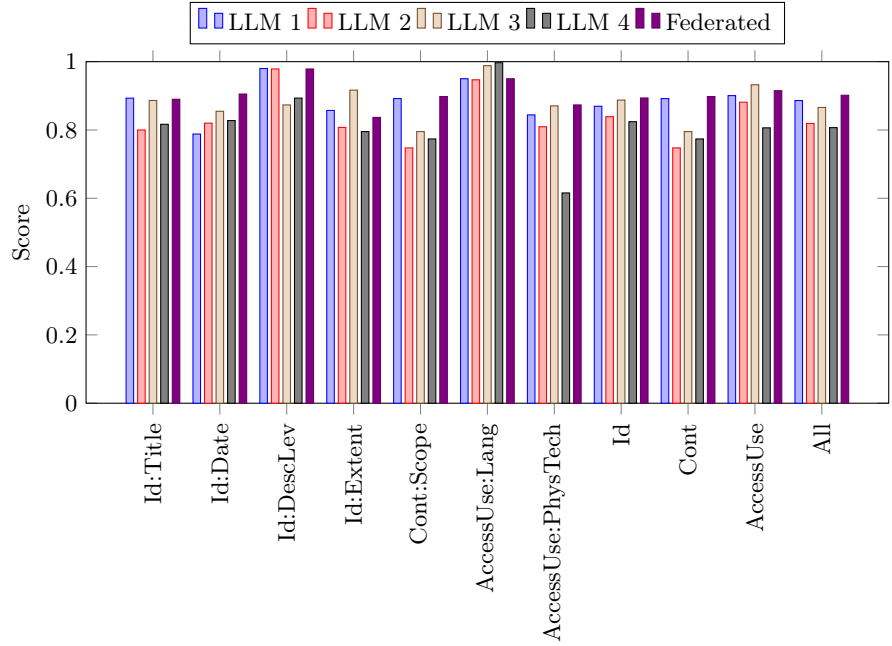

\begin{table}[h!]
\caption{Improvements of the LLM scores of the federated approach in comparison to direct use of LLMs}\label{tab:Improvements}
\centering
\scriptsize
\begin{tabular}{|l | r | r | r | r|}
    \hline 
    \textbf{Element/Area} & \textbf{LLM 1} & \textbf{LLM 2} & \textbf{LLM 3} & \textbf{LLM 4} \\
    \hline\hline
Id:Title & 0\% & 11\% & 0\% & 9\% \\
Id:Date & 15\% & 10\% & 6\% & 9\% \\
Id:DescLev & 0\% & 0\% & 12\% & 10\% \\
Id:Extent & -2\% & 4\% & -9\% & 5\% \\
Cont:Scope & 1\% & 20\% & 13\% & 16\% \\
AccessUse:Lang & 0\% & 0\% & -4\% & -5\% \\
AccessUse:PhysTech & 3\% & 8\% & 0\% & 42\% \\
\hline
Id & 3\% & 7\% & 1\% & 8\% \\
Cont & 1\% & 20\% & 13\% & 16\% \\
AccessUse & 2\% & 4\% & -2\% & 13\% \\
\hline
All & 2\% & 10\% & 4\% & 12\% \\
\hline
\end{tabular}
\end{table}

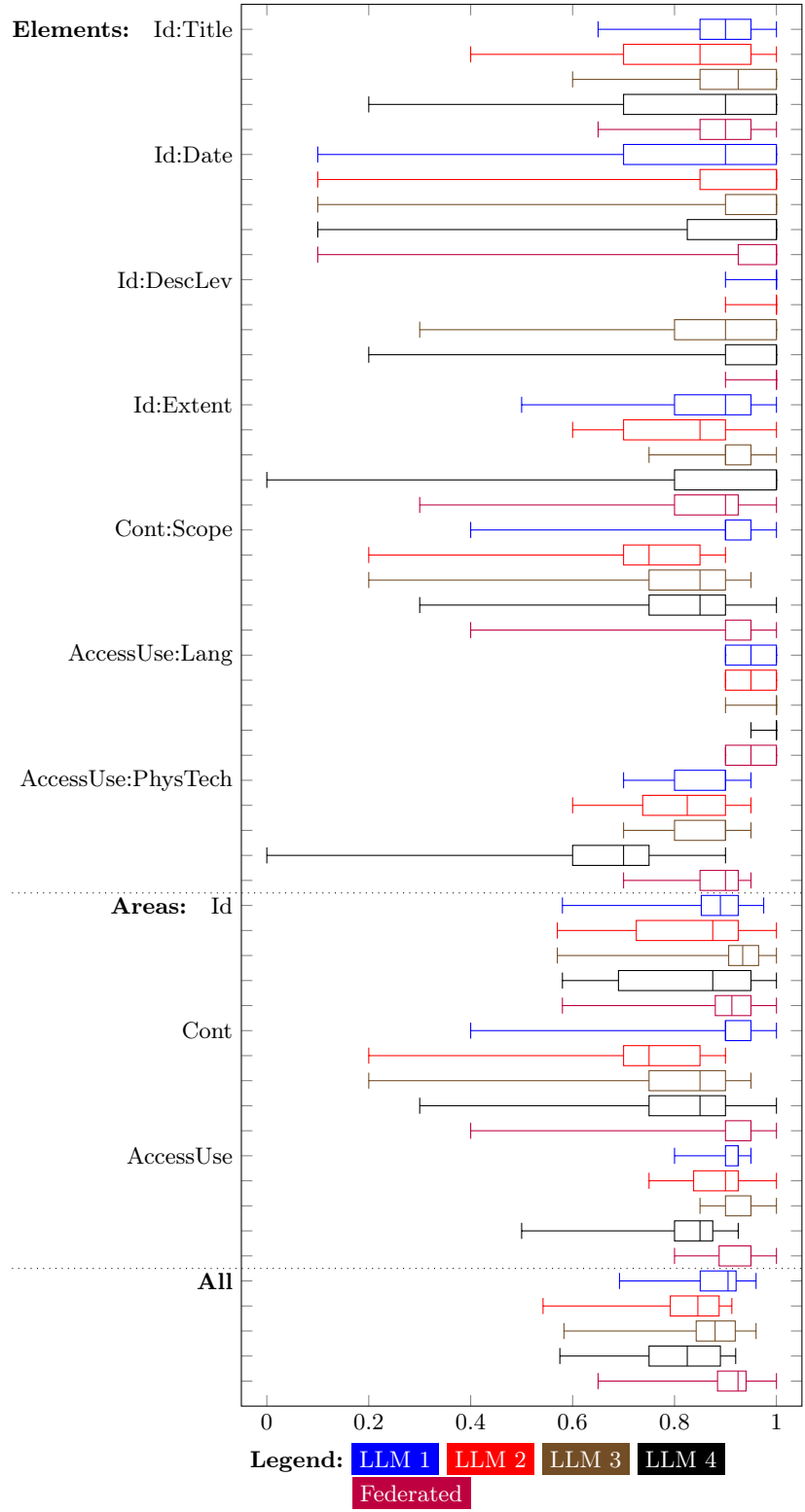
\begin{figure}
\begin{tikzpicture}
\begin{axis}
[
    ytick={1, 2, 3, 4, 5, 6, 7, 8, 9, 10, 11, 12, 13, 14, 15, 16, 17, 18, 19, 20, 21, 22, 23, 24, 25, 26, 27, 28, 29, 30, 31, 32, 33, 34, 35, 36, 37, 38, 39, 40, 41, 42, 43, 44, 45, 46, 47, 48, 49, 50, 51, 52, 53, 54, 55},
    yticklabels={, , , , \textbf{All}, , , , , AccessUse, , , , , Cont, , , , , \textbf{Areas:}\quad Id, , , , , AccessUse:PhysTech, , , , , AccessUse:Lang, , , , , Cont:Scope, , , , , Id:Extent, , , , , Id:DescLev, , , , , Id:Date, , , , , \textbf{Elements:}\quad Id:Title},
    y post scale=3,
    width=\textwidth-3cm,
    ymin=0,
    ymax=56,
    xmin=-0.05,
    xmax=1.05,
    clip=false
]
\addplot+[draw=purple,solid,
    boxplot prepared={
        median=0.9246000000000001,
        upper quartile=0.940425,
        lower quartile=0.8843749999999999,
        upper whisker=1.0,
        lower whisker=0.65
    },
    ] coordinates {};
\addplot+[draw=black,solid,
    boxplot prepared={
        median=0.825,
        upper quartile=0.89,
        lower quartile=0.75,
        upper whisker=0.92,
        lower whisker=0.575
    },
    ] coordinates {};
\addplot+[draw=brown!60!black,solid,
    boxplot prepared={
        median=0.8795999999999999,
        upper quartile=0.9191750000000001,
        lower quartile=0.8425,
        upper whisker=0.96,
        lower whisker=0.583
    },
    ] coordinates {};
\addplot+[draw=red,solid,
    boxplot prepared={
        median=0.8458,
        upper quartile=0.8875,
        lower quartile=0.7917,
        upper whisker=0.9125,
        lower whisker=0.542
    },
    ] coordinates {};
\addplot+[draw=blue,solid,
    boxplot prepared={
        median=0.905,
        upper quartile=0.9208,
        lower quartile=0.850375,
        upper whisker=0.96,
        lower whisker=0.692
    },
    ] coordinates {};
\addplot+[draw=purple,solid,
    boxplot prepared={
        median=0.95,
        upper quartile=0.95,
        lower quartile=0.8875,
        upper whisker=1.0,
        lower whisker=0.8
    },
    ] coordinates {};
\addplot+[draw=black,solid,
    boxplot prepared={
        median=0.85,
        upper quartile=0.875,
        lower quartile=0.8,
        upper whisker=0.925,
        lower whisker=0.5
    },
    ] coordinates {};
\addplot+[draw=brown!60!black,solid,
    boxplot prepared={
        median=0.95,
        upper quartile=0.95,
        lower quartile=0.9,
        upper whisker=1.0,
        lower whisker=0.85
    },
    ] coordinates {};
\addplot+[draw=red,solid,
    boxplot prepared={
        median=0.9,
        upper quartile=0.925,
        lower quartile=0.8375,
        upper whisker=1.0,
        lower whisker=0.75
    },
    ] coordinates {};
\addplot+[draw=blue,solid,
    boxplot prepared={
        median=0.925,
        upper quartile=0.925,
        lower quartile=0.9,
        upper whisker=0.95,
        lower whisker=0.8
    },
    ] coordinates {};
\addplot+[draw=purple,solid,
    boxplot prepared={
        median=0.95,
        upper quartile=0.95,
        lower quartile=0.9,
        upper whisker=1.0,
        lower whisker=0.4
    },
    ] coordinates {};
\addplot+[draw=black,solid,
    boxplot prepared={
        median=0.85,
        upper quartile=0.9,
        lower quartile=0.75,
        upper whisker=1.0,
        lower whisker=0.3
    },
    ] coordinates {};
\addplot+[draw=brown!60!black,solid,
    boxplot prepared={
        median=0.85,
        upper quartile=0.9,
        lower quartile=0.75,
        upper whisker=0.95,
        lower whisker=0.2
    },
    ] coordinates {};
\addplot+[draw=red,solid,
    boxplot prepared={
        median=0.75,
        upper quartile=0.85,
        lower quartile=0.7,
        upper whisker=0.9,
        lower whisker=0.2
    },
    ] coordinates {};
\addplot+[draw=blue,solid,
    boxplot prepared={
        median=0.9,
        upper quartile=0.95,
        lower quartile=0.9,
        upper whisker=1.0,
        lower whisker=0.4
    },
    ] coordinates {};
\addplot+[draw=purple,solid,
    boxplot prepared={
        median=0.9125000000000001,
        upper quartile=0.95,
        lower quartile=0.88,
        upper whisker=1.0,
        lower whisker=0.58
    },
    ] coordinates {};
\addplot+[draw=black,solid,
    boxplot prepared={
        median=0.875,
        upper quartile=0.95,
        lower quartile=0.69,
        upper whisker=1.0,
        lower whisker=0.58
    },
    ] coordinates {};
\addplot+[draw=brown!60!black,solid,
    boxplot prepared={
        median=0.9337500000000001,
        upper quartile=0.965,
        lower quartile=0.90625,
        upper whisker=1.0,
        lower whisker=0.57
    },
    ] coordinates {};
\addplot+[draw=red,solid,
    boxplot prepared={
        median=0.875,
        upper quartile=0.925,
        lower quartile=0.725,
        upper whisker=1.0,
        lower whisker=0.57
    },
    ] coordinates {};
\addplot+[draw=blue,solid,
    boxplot prepared={
        median=0.89,
        upper quartile=0.925,
        lower quartile=0.8525,
        upper whisker=0.975,
        lower whisker=0.58
    },
    ] coordinates {};
\addplot+[draw=purple,solid,
    boxplot prepared={
        median=0.9,
        upper quartile=0.925,
        lower quartile=0.8500000000000001,
        upper whisker=0.95,
        lower whisker=0.7
    },
    ] coordinates {};
\addplot+[draw=black,solid,
    boxplot prepared={
        median=0.7,
        upper quartile=0.75,
        lower quartile=0.6,
        upper whisker=0.9,
        lower whisker=0.0
    },
    ] coordinates {};
\addplot+[draw=brown!60!black,solid,
    boxplot prepared={
        median=0.9,
        upper quartile=0.9,
        lower quartile=0.8,
        upper whisker=0.95,
        lower whisker=0.7
    },
    ] coordinates {};
\addplot+[draw=red,solid,
    boxplot prepared={
        median=0.825,
        upper quartile=0.9,
        lower quartile=0.7375,
        upper whisker=0.95,
        lower whisker=0.6
    },
    ] coordinates {};
\addplot+[draw=blue,solid,
    boxplot prepared={
        median=0.9,
        upper quartile=0.9,
        lower quartile=0.8,
        upper whisker=0.95,
        lower whisker=0.7
    },
    ] coordinates {};
\addplot+[draw=purple,solid,
    boxplot prepared={
        median=0.95,
        upper quartile=1.0,
        lower quartile=0.9,
        upper whisker=1.0,
        lower whisker=0.9
    },
    ] coordinates {};
\addplot+[draw=black,solid,
    boxplot prepared={
        median=1.0,
        upper quartile=1.0,
        lower quartile=1.0,
        upper whisker=1.0,
        lower whisker=0.95
    },
    ] coordinates {};
\addplot+[draw=brown!60!black,solid,
    boxplot prepared={
        median=1.0,
        upper quartile=1.0,
        lower quartile=1.0,
        upper whisker=1.0,
        lower whisker=0.9
    },
    ] coordinates {};
\addplot+[draw=red,solid,
    boxplot prepared={
        median=0.95,
        upper quartile=1.0,
        lower quartile=0.9,
        upper whisker=1.0,
        lower whisker=0.9
    },
    ] coordinates {};
\addplot+[draw=blue,solid,
    boxplot prepared={
        median=0.95,
        upper quartile=1.0,
        lower quartile=0.9,
        upper whisker=1.0,
        lower whisker=0.9
    },
    ] coordinates {};
\addplot+[draw=purple,solid,
    boxplot prepared={
        median=0.95,
        upper quartile=0.95,
        lower quartile=0.9,
        upper whisker=1.0,
        lower whisker=0.4
    },
    ] coordinates {};
\addplot+[draw=black,solid,
    boxplot prepared={
        median=0.85,
        upper quartile=0.9,
        lower quartile=0.75,
        upper whisker=1.0,
        lower whisker=0.3
    },
    ] coordinates {};
\addplot+[draw=brown!60!black,solid,
    boxplot prepared={
        median=0.85,
        upper quartile=0.9,
        lower quartile=0.75,
        upper whisker=0.95,
        lower whisker=0.2
    },
    ] coordinates {};
\addplot+[draw=red,solid,
    boxplot prepared={
        median=0.75,
        upper quartile=0.85,
        lower quartile=0.7,
        upper whisker=0.9,
        lower whisker=0.2
    },
    ] coordinates {};
\addplot+[draw=blue,solid,
    boxplot prepared={
        median=0.9,
        upper quartile=0.95,
        lower quartile=0.9,
        upper whisker=1.0,
        lower whisker=0.4
    },
    ] coordinates {};
\addplot+[draw=purple,solid,
    boxplot prepared={
        median=0.9,
        upper quartile=0.925,
        lower quartile=0.8,
        upper whisker=1.0,
        lower whisker=0.3
    },
    ] coordinates {};
\addplot+[draw=black,solid,
    boxplot prepared={
        median=1.0,
        upper quartile=1.0,
        lower quartile=0.8,
        upper whisker=1.0,
        lower whisker=0.0
    },
    ] coordinates {};
\addplot+[draw=brown!60!black,solid,
    boxplot prepared={
        median=0.9,
        upper quartile=0.95,
        lower quartile=0.9,
        upper whisker=1.0,
        lower whisker=0.75
    },
    ] coordinates {};
\addplot+[draw=red,solid,
    boxplot prepared={
        median=0.8500000000000001,
        upper quartile=0.9,
        lower quartile=0.7,
        upper whisker=1.0,
        lower whisker=0.6
    },
    ] coordinates {};
\addplot+[draw=blue,solid,
    boxplot prepared={
        median=0.9,
        upper quartile=0.95,
        lower quartile=0.8,
        upper whisker=1.0,
        lower whisker=0.5
    },
    ] coordinates {};
\addplot+[draw=purple,solid,
    boxplot prepared={
        median=1.0,
        upper quartile=1.0,
        lower quartile=1.0,
        upper whisker=1.0,
        lower whisker=0.9
    },
    ] coordinates {};
\addplot+[draw=black,solid,
    boxplot prepared={
        median=1.0,
        upper quartile=1.0,
        lower quartile=0.9,
        upper whisker=1.0,
        lower whisker=0.2
    },
    ] coordinates {};
\addplot+[draw=brown!60!black,solid,
    boxplot prepared={
        median=0.9,
        upper quartile=1.0,
        lower quartile=0.8,
        upper whisker=1.0,
        lower whisker=0.3
    },
    ] coordinates {};
\addplot+[draw=red,solid,
    boxplot prepared={
        median=1.0,
        upper quartile=1.0,
        lower quartile=1.0,
        upper whisker=1.0,
        lower whisker=0.9
    },
    ] coordinates {};
\addplot+[draw=blue,solid,
    boxplot prepared={
        median=1.0,
        upper quartile=1.0,
        lower quartile=1.0,
        upper whisker=1.0,
        lower whisker=0.9
    },
    ] coordinates {};
\addplot+[draw=purple,solid,
    boxplot prepared={
        median=1.0,
        upper quartile=1.0,
        lower quartile=0.925,
        upper whisker=1.0,
        lower whisker=0.1
    },
    ] coordinates {};
\addplot+[draw=black,solid,
    boxplot prepared={
        median=1.0,
        upper quartile=1.0,
        lower quartile=0.825,
        upper whisker=1.0,
        lower whisker=0.1
    },
    ] coordinates {};
\addplot+[draw=brown!60!black,solid,
    boxplot prepared={
        median=1.0,
        upper quartile=1.0,
        lower quartile=0.9,
        upper whisker=1.0,
        lower whisker=0.1
    },
    ] coordinates {};
\addplot+[draw=red,solid,
    boxplot prepared={
        median=1.0,
        upper quartile=1.0,
        lower quartile=0.85,
        upper whisker=1.0,
        lower whisker=0.1
    },
    ] coordinates {};
\addplot+[draw=blue,solid,
    boxplot prepared={
        median=0.9,
        upper quartile=1.0,
        lower quartile=0.7,
        upper whisker=1.0,
        lower whisker=0.1
    },
    ] coordinates {};
\addplot+[draw=purple,solid,
    boxplot prepared={
        median=0.9,
        upper quartile=0.95,
        lower quartile=0.85,
        upper whisker=1.0,
        lower whisker=0.65
    },
    ] coordinates {};
\addplot+[draw=black,solid,
    boxplot prepared={
        median=0.9,
        upper quartile=1.0,
        lower quartile=0.7,
        upper whisker=1.0,
        lower whisker=0.2
    },
    ] coordinates {};
\addplot+[draw=brown!60!black,solid,
    boxplot prepared={
        median=0.925,
        upper quartile=1.0,
        lower quartile=0.85,
        upper whisker=1.0,
        lower whisker=0.6
    },
    ] coordinates {};
\addplot+[draw=red,solid,
    boxplot prepared={
        median=0.85,
        upper quartile=0.95,
        lower quartile=0.7,
        upper whisker=1.0,
        lower whisker=0.4
    },
    ] coordinates {};
\addplot+[draw=blue,solid,
    boxplot prepared={
        median=0.9,
        upper quartile=0.95,
        lower quartile=0.85,
        upper whisker=1.0,
        lower whisker=0.65
    },
    ] coordinates {};
    \node[below right, align=left, yshift=-1.2em] {\textbf{Legend:} \colorbox{blue}{\color{white}LLM 1} \colorbox{red}{\color{white}LLM 2}  \colorbox{brown!60!black}{\color{white}LLM 3} \colorbox{black}{\color{white}LLM 4}\\\hspace{4.28em}\colorbox{purple}{\color{white}Federated}};
    \draw[dotted] (-0.5,5.5) -- (1.05,5.5);
    \draw[dotted] (-0.5,20.5) -- (1.05,20.5);

\end{axis}
\end{tikzpicture}
\caption{Box plots of LLM scores for information extraction}\label{fig:boxplots}
\end{figure}

The ground truth was manually created by professional archivists of the AdsD and includes a full set of ISAD(G) metadata for archival materials in the dataset.  ISAD(G) stands for General International Standard Archival Description, a framework developed by the International Council on Archives (ICA) to standardize the description of archival materials. 
It is widely used by archivists globally to make archival records more accessible and understandable. 
ISAD(G) organizes metadata elements into several areas, and each element serves a specific purpose in creating a comprehensive and standardized archival description. These elements can be divided into two groups: one group of elements whose values can be determined directly or by inferring from the archive material itself; another group of elements whose values need information from external sources or are assigned by an archivist. As a result, the elements in the second group are not suitable for evaluation.  We list the areas and elements of ISAD(G) considered in our evaluation in Table~\ref{tab:ISADG}, and also introduce their one-word tags,  which we use in the succeeding figures (Fig.~\ref{fig:totaleval} and Fig.~\ref{fig:boxplots}) and tables (Table~\ref{tab:Improvements}).

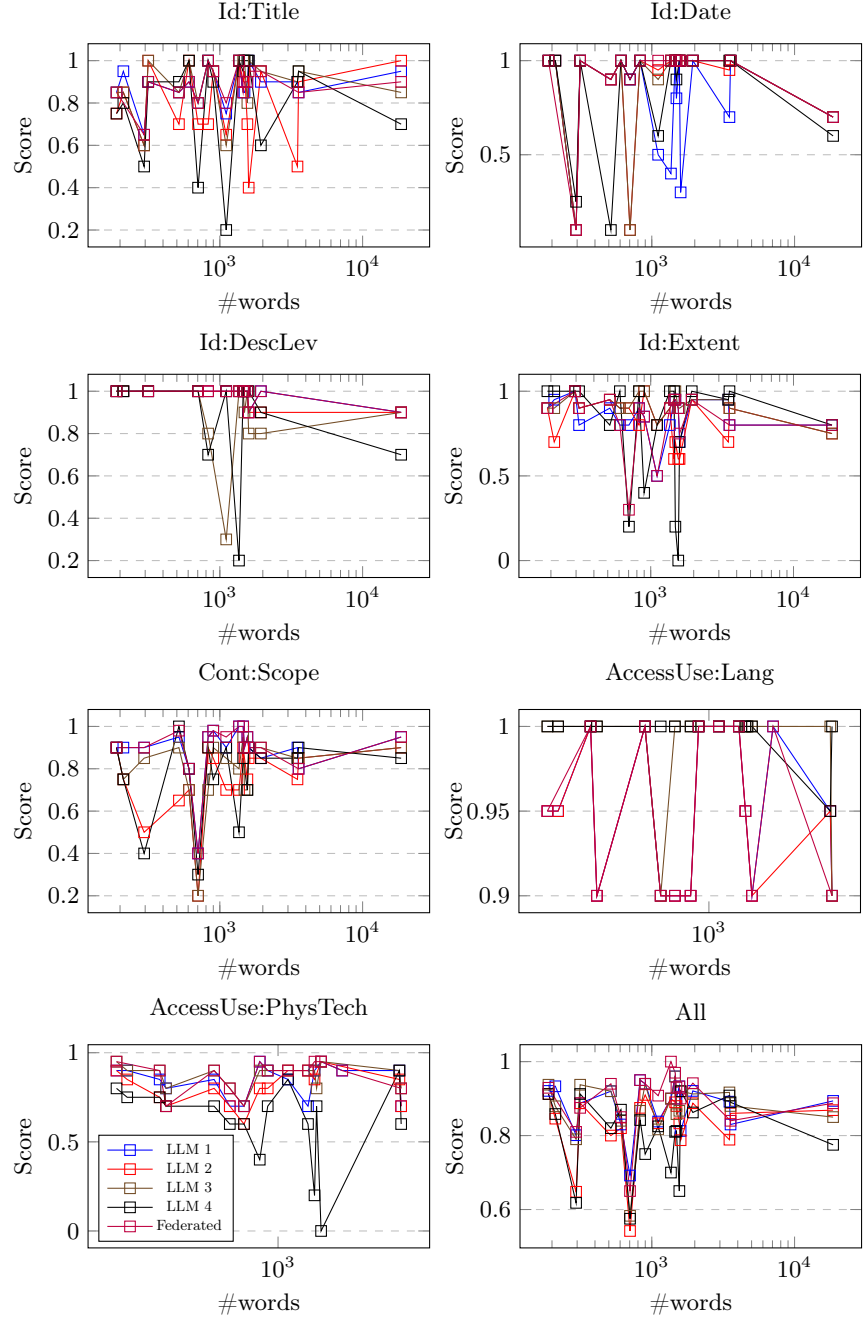
\begin{figure}
\begin{tikzpicture}
\begin{axis}[
    title={Id:Title},
    xlabel={\#words},
    ylabel={Score},
    legend pos=south east,
    ymajorgrids=true,
    grid style=dashed,
    width=.5\textwidth,
    height=.35\textwidth,
    xmode=log
]
\addplot[
    color=blue,
    mark=square,
    ]
    coordinates {
    (189,0.85)
    (211,0.95)
    (295,0.65)
    (315,0.9)
    (517,0.85)
    (608,0.9)
    (704,0.8)
    (828,1.0)
    (899,0.95)
    (1108,0.75)
    (1362,1.0)
    (1457,0.85)
    (1488,0.85)
    (1559,0.9)
    (1592,1.0)
    (1939,0.9)
    (3497,0.9)
    (3561,0.85)
    (18616,0.95)
    };
    \addlegendentry{LLM 1}
\addplot[
    color=red,
    mark=square,
    ]
    coordinates {
    (189,0.75)
    (211,0.85)
    (295,0.6)
    (315,1.0)
    (517,0.7)
    (608,1.0)
    (704,0.7)
    (828,0.7)
    (899,0.95)
    (1108,0.65)
    (1362,1.0)
    (1457,0.9)
    (1488,0.95)
    (1559,0.7)
    (1592,0.4)
    (1939,0.95)
    (3497,0.5)
    (3561,0.9)
    (18616,1.0)
    };
    \addlegendentry{LLM 2}
\addplot[
    color=brown!60!black,
    mark=square,
    ]
    coordinates {
    (189,0.85)
    (211,0.85)
    (295,0.6)
    (315,1.0)
    (517,0.85)
    (608,1.0)
    (704,0.8)
    (828,1.0)
    (899,0.95)
    (1108,0.6)
    (1362,1.0)
    (1457,1.0)
    (1488,1.0)
    (1559,0.8)
    (1592,1.0)
    (1939,0.95)
    (3497,0.9)
    (3561,0.95)
    (18616,0.85)
    };
    \addlegendentry{LLM 3}
\addplot[
    color=black,
    mark=square,
    ]
    coordinates {
    (189,0.75)
    (211,0.8)
    (295,0.5)
    (315,0.9)
    (517,0.9)
    (608,1.0)
    (704,0.4)
    (828,1.0)
    (899,0.9)
    (1108,0.2)
    (1362,1.0)
    (1457,0.95)
    (1488,1.0)
    (1559,1.0)
    (1592,1.0)
    (1939,0.6)
    (3497,0.9)
    (3561,0.95)
    (18616,0.7)
    };
    \addlegendentry{LLM 4}
\addplot[
    color=purple,
    mark=square,
    ]
    coordinates {
    (189,0.85)
    (295,0.65)
    (315,0.9)
    (517,0.85)
    (608,0.9)
    (704,0.8)
    (828,1.0)
    (899,0.95)
    (1108,0.8)
    (1362,1.0)
    (1457,0.85)
    (1488,0.9)
    (1559,0.9)
    (1592,0.95)
    (1939,0.95)
    (3561,0.85)
    (18616,0.9)
    };
    \addlegendentry{Federated}
    \legend{}; 
\end{axis}
\end{tikzpicture}
\begin{tikzpicture}
\begin{axis}[
    title={Id:Date},
    xlabel={\#words},
    ylabel={Score},
    legend pos=south east,
    ymajorgrids=true,
    grid style=dashed,
    width=.5\textwidth,
    height=.35\textwidth,
    xmode=log
]
\addplot[
    color=blue,
    mark=square,
    ]
    coordinates {
    (189,1.0)
    (211,1.0)
    (295,0.1)
    (315,1.0)
    (517,0.9)
    (608,1.0)
    (704,0.9)
    (828,1.0)
    (1108,0.5)
    (1362,0.4)
    (1457,1.0)
    (1488,0.8)
    (1559,1.0)
    (1592,0.3)
    (1939,1.0)
    (3497,0.7)
    (3561,1.0)
    (18616,0.7)
    };
    \addlegendentry{LLM 1}
\addplot[
    color=red,
    mark=square,
    ]
    coordinates {
    (189,1.0)
    (211,1.0)
    (295,0.1)
    (315,1.0)
    (517,0.9)
    (608,1.0)
    (704,0.1)
    (828,1.0)
    (1108,0.95)
    (1362,1.0)
    (1457,1.0)
    (1488,1.0)
    (1559,1.0)
    (1592,1.0)
    (1939,1.0)
    (3497,0.95)
    (3561,1.0)
    (18616,0.7)
    };
    \addlegendentry{LLM 2}
\addplot[
    color=brown!60!black,
    mark=square,
    ]
    coordinates {
    (189,1.0)
    (211,1.0)
    (295,0.1)
    (315,1.0)
    (517,0.9)
    (608,1.0)
    (704,0.1)
    (828,1.0)
    (1108,0.9)
    (1362,1.0)
    (1457,1.0)
    (1488,1.0)
    (1559,1.0)
    (1592,1.0)
    (1939,1.0)
    (3497,1.0)
    (3561,1.0)
    (18616,0.7)
    };
    \addlegendentry{LLM 3}
\addplot[
    color=black,
    mark=square,
    ]
    coordinates {
    (189,1.0)
    (211,1.0)
    (295,0.25)
    (315,1.0)
    (517,0.1)
    (608,1.0)
    (704,0.9)
    (828,1.0)
    (1108,0.6)
    (1362,1.0)
    (1457,1.0)
    (1488,0.9)
    (1559,1.0)
    (1592,1.0)
    (1939,1.0)
    (3497,1.0)
    (3561,1.0)
    (18616,0.6)
    };
    \addlegendentry{LLM 4}
\addplot[
    color=purple,
    mark=square,
    ]
    coordinates {
    (189,1.0)
    (295,0.1)
    (315,1.0)
    (517,0.9)
    (608,1.0)
    (704,0.9)
    (828,1.0)
    (1108,1.0)
    (1362,1.0)
    (1457,1.0)
    (1488,1.0)
    (1559,1.0)
    (1592,1.0)
    (1939,1.0)
    (3561,1.0)
    (18616,0.7)
    };
    \addlegendentry{Federated}
    \legend{}; 
\end{axis}
\end{tikzpicture}\\
\begin{tikzpicture}
\begin{axis}[
    title={Id:DescLev},
    xlabel={\#words},
    ylabel={Score},
    legend pos=south east,
    ymajorgrids=true,
    grid style=dashed,
    width=.5\textwidth,
    height=.35\textwidth,
    xmode=log
]
\addplot[
    color=blue,
    mark=square,
    ]
    coordinates {
    (189,1.0)
    (211,1.0)
    (315,1.0)
    (704,1.0)
    (828,1.0)
    (1108,1.0)
    (1362,1.0)
    (1457,1.0)
    (1488,1.0)
    (1559,1.0)
    (1592,0.9)
    (1939,1.0)
    (18616,0.9)
    };
    \addlegendentry{LLM 1}
\addplot[
    color=red,
    mark=square,
    ]
    coordinates {
    (189,1.0)
    (211,1.0)
    (315,1.0)
    (704,1.0)
    (828,1.0)
    (1108,1.0)
    (1362,1.0)
    (1457,1.0)
    (1488,1.0)
    (1559,1.0)
    (1592,0.9)
    (1939,0.9)
    (18616,0.9)
    };
    \addlegendentry{LLM 2}
\addplot[
    color=brown!60!black,
    mark=square,
    ]
    coordinates {
    (189,1.0)
    (211,1.0)
    (315,1.0)
    (704,1.0)
    (828,0.8)
    (1108,0.3)
    (1362,1.0)
    (1457,1.0)
    (1488,0.9)
    (1559,1.0)
    (1592,0.8)
    (1939,0.8)
    (18616,0.9)
    };
    \addlegendentry{LLM 3}
\addplot[
    color=black,
    mark=square,
    ]
    coordinates {
    (189,1.0)
    (211,1.0)
    (315,1.0)
    (704,1.0)
    (828,0.7)
    (1108,1.0)
    (1362,0.2)
    (1457,1.0)
    (1488,1.0)
    (1559,1.0)
    (1592,1.0)
    (1939,0.9)
    (18616,0.7)
    };
    \addlegendentry{LLM 4}
\addplot[
    color=purple,
    mark=square,
    ]
    coordinates {
    (189,1.0)
    (315,1.0)
    (704,1.0)
    (828,1.0)
    (1108,1.0)
    (1362,1.0)
    (1457,1.0)
    (1488,1.0)
    (1559,1.0)
    (1592,0.9)
    (1939,1.0)
    (18616,0.9)
    };
    \addlegendentry{Federated}
    \legend{}; 
\end{axis}
\end{tikzpicture}
\begin{tikzpicture}
\hspace{-2.75em}
\begin{axis}[
    title={Id:Extent},
    xlabel={\#words},
    ylabel={Score},
    legend pos=south east,
    ymajorgrids=true,
    grid style=dashed,
    width=.5\textwidth,
    height=.35\textwidth,
    xmode=log
]
\addplot[
    color=blue,
    mark=square,
    ]
    coordinates {
    (189,0.9)
    (211,0.95)
    (295,1.0)
    (315,0.8)
    (517,0.9)
    (608,0.8)
    (704,0.8)
    (828,0.9)
    (899,0.85)
    (1108,0.5)
    (1362,0.8)
    (1457,0.95)
    (1488,0.95)
    (1559,0.9)
    (1592,0.7)
    (1939,0.95)
    (3497,0.95)
    (3561,0.8)
    (18616,0.8)
    };
    \addlegendentry{LLM 1}
\addplot[
    color=red,
    mark=square,
    ]
    coordinates {
    (189,0.9)
    (211,0.7)
    (295,1.0)
    (315,0.9)
    (517,0.95)
    (608,0.9)
    (704,0.9)
    (828,0.8)
    (899,1.0)
    (1108,0.8)
    (1362,0.9)
    (1457,0.6)
    (1488,0.7)
    (1559,0.6)
    (1592,0.6)
    (1939,0.95)
    (3497,0.7)
    (3561,0.9)
    (18616,0.75)
    };
    \addlegendentry{LLM 2}
\addplot[
    color=brown!60!black,
    mark=square,
    ]
    coordinates {
    (189,0.9)
    (211,0.9)
    (295,1.0)
    (315,0.9)
    (517,0.95)
    (608,0.9)
    (704,0.9)
    (828,1.0)
    (899,1.0)
    (1108,0.8)
    (1362,0.9)
    (1457,0.95)
    (1488,1.0)
    (1559,0.9)
    (1592,0.9)
    (1939,0.95)
    (3497,0.95)
    (3561,0.9)
    (18616,0.75)
    };
    \addlegendentry{LLM 3}
\addplot[
    color=black,
    mark=square,
    ]
    coordinates {
    (189,1.0)
    (211,1.0)
    (295,1.0)
    (315,1.0)
    (517,0.8)
    (608,1.0)
    (704,0.2)
    (828,1.0)
    (899,0.4)
    (1108,0.8)
    (1362,1.0)
    (1457,1.0)
    (1488,0.2)
    (1559,0.0)
    (1592,0.7)
    (1939,1.0)
    (3497,0.95)
    (3561,1.0)
    (18616,0.8)
    };
    \addlegendentry{LLM 4}
\addplot[
    color=purple,
    mark=square,
    ]
    coordinates {
    (189,0.9)
    (295,1.0)
    (315,0.9)
    (517,0.95)
    (608,0.8)
    (704,0.3)
    (828,0.9)
    (899,0.85)
    (1108,0.5)
    (1362,0.9)
    (1457,0.95)
    (1488,0.95)
    (1559,0.9)
    (1592,0.75)
    (1939,0.95)
    (3561,0.8)
    (18616,0.8)
    };
    \addlegendentry{Federated}
    \legend{}; 
\end{axis}
\end{tikzpicture}
\begin{tikzpicture}
\begin{axis}[
    title={Cont:Scope},
    xlabel={\#words},
    ylabel={Score},
    legend pos=south east,
    ymajorgrids=true,
    grid style=dashed,
    width=.5\textwidth,
    height=.35\textwidth,
    xmode=log
]
\addplot[
    color=blue,
    mark=square,
    ]
    coordinates {
    (189,0.9)
    (211,0.9)
    (295,0.9)
    (517,0.95)
    (608,0.8)
    (704,0.4)
    (828,0.95)
    (899,0.98)
    (1108,0.9)
    (1362,1.0)
    (1457,1.0)
    (1488,0.9)
    (1559,0.95)
    (1592,0.9)
    (1939,0.85)
    (3497,0.9)
    (3561,0.8)
    (18616,0.95)
    };
    \addlegendentry{LLM 1}
\addplot[
    color=red,
    mark=square,
    ]
    coordinates {
    (189,0.9)
    (211,0.75)
    (295,0.5)
    (517,0.65)
    (608,0.7)
    (704,0.2)
    (828,0.9)
    (899,0.85)
    (1108,0.7)
    (1362,0.7)
    (1457,0.9)
    (1488,0.85)
    (1559,0.75)
    (1592,0.85)
    (1939,0.85)
    (3497,0.75)
    (3561,0.85)
    (18616,0.9)
    };
    \addlegendentry{LLM 2}
\addplot[
    color=brown!60!black,
    mark=square,
    ]
    coordinates {
    (189,0.9)
    (211,0.75)
    (295,0.85)
    (517,0.9)
    (608,0.7)
    (704,0.2)
    (828,0.7)
    (899,0.9)
    (1108,0.85)
    (1362,0.8)
    (1457,0.95)
    (1488,0.7)
    (1559,0.7)
    (1592,0.9)
    (1939,0.9)
    (3497,0.85)
    (3561,0.85)
    (18616,0.9)
    };
    \addlegendentry{LLM 3}
\addplot[
    color=black,
    mark=square,
    ]
    coordinates {
    (189,0.9)
    (211,0.75)
    (295,0.4)
    (517,1.0)
    (608,0.8)
    (704,0.3)
    (828,0.9)
    (899,0.75)
    (1108,0.9)
    (1362,0.5)
    (1457,0.85)
    (1559,0.7)
    (1592,0.9)
    (1939,0.85)
    (3497,0.85)
    (3561,0.9)
    (18616,0.85)
    };
    \addlegendentry{LLM 4}
\addplot[
    color=purple,
    mark=square,
    ]
    coordinates {
    (189,0.9)
    (295,0.9)
    (517,0.98)
    (608,0.8)
    (704,0.4)
    (828,0.95)
    (899,0.98)
    (1108,0.95)
    (1362,1.0)
    (1457,1.0)
    (1488,0.9)
    (1559,0.95)
    (1592,0.9)
    (1939,0.9)
    (3561,0.8)
    (18616,0.95)
    };
    \addlegendentry{Federated}
    \legend{}; 
\end{axis}
\end{tikzpicture}
\begin{tikzpicture}
\hspace{-2.75em}
\begin{axis}[
    title={AccessUse:Lang},
    xlabel={\#words},
    ylabel={Score},
    legend pos=south east,
    ymajorgrids=true,
    grid style=dashed,
    width=.5\textwidth,
    height=.35\textwidth,
    xmode=log,
    ylabel shift = {-0.5em}
]
\addplot[
    color=blue,
    mark=square,
    ]
    coordinates {
    (189,0.95)
    (211,0.95)
    (295,1.0)
    (315,0.9)
    (517,1.0)
    (608,0.9)
    (704,0.9)
    (828,0.9)
    (899,1.0)
    (1108,1.0)
    (1362,1.0)
    (1457,0.95)
    (1559,0.9)
    (1939,1.0)
    (3497,0.95)
    (3561,0.9)
    };
    \addlegendentry{LLM 1}
\addplot[
    color=red,
    mark=square,
    ]
    coordinates {
    (189,0.95)
    (211,0.95)
    (295,1.0)
    (315,0.9)
    (517,1.0)
    (608,0.9)
    (704,0.9)
    (828,0.9)
    (899,1.0)
    (1108,1.0)
    (1362,1.0)
    (1457,0.95)
    (1559,0.9)
    (3497,0.95)
    (3561,0.9)
    };
    \addlegendentry{LLM 2}
\addplot[
    color=brown!60!black,
    mark=square,
    ]
    coordinates {
    (189,1.0)
    (211,1.0)
    (295,1.0)
    (315,1.0)
    (517,1.0)
    (608,0.9)
    (704,1.0)
    (828,1.0)
    (899,1.0)
    (1108,1.0)
    (1362,1.0)
    (1457,1.0)
    (1488,1.0)
    (1559,1.0)
    (3497,1.0)
    (3561,0.9)
    };
    \addlegendentry{LLM 3}
\addplot[
    color=black,
    mark=square,
    ]
    coordinates {
    (189,1.0)
    (211,1.0)
    (295,1.0)
    (315,1.0)
    (517,1.0)
    (608,1.0)
    (704,1.0)
    (828,1.0)
    (899,1.0)
    (1108,1.0)
    (1362,1.0)
    (1457,1.0)
    (1488,1.0)
    (1559,1.0)
    (3497,0.95)
    (3561,1.0)
    };
    \addlegendentry{LLM 4}
\addplot[
    color=purple,
    mark=square,
    ]
    coordinates {
    (189,0.95)
    (295,1.0)
    (315,0.9)
    (517,1.0)
    (608,0.9)
    (704,0.9)
    (828,0.9)
    (899,1.0)
    (1108,1.0)
    (1362,1.0)
    (1457,0.95)
    (1559,0.9)
    (1939,1.0)
    (3561,0.9)
    };
    \addlegendentry{Federated}
    \legend{}; 
\end{axis}
\end{tikzpicture}
\begin{tikzpicture}
\begin{axis}[
    title={AccessUse:PhysTech},
    xlabel={\#words},
    ylabel={Score},
    legend pos=south west,
    ymajorgrids=true,
    grid style=dashed,
    width=.5\textwidth,
    height=.35\textwidth,
    xmode=log,
    legend style={nodes={scale=0.62, transform shape}}
]
\addplot[
    color=blue,
    mark=square,
    ]
    coordinates {
    (189,0.9)
    (211,0.9)
    (295,0.85)
    (315,0.8)
    (517,0.85)
    (608,0.7)
    (704,0.7)
    (828,0.95)
    (899,0.9)
    (1108,0.85)
    (1362,0.7)
    (1457,0.9)
    (1559,0.95)
    (1939,0.9)
    (3497,0.9)
    (3561,0.7)
    };
    \addlegendentry{LLM 1}
\addplot[
    color=red,
    mark=square,
    ]
    coordinates {
    (189,0.9)
    (211,0.85)
    (295,0.75)
    (315,0.7)
    (517,0.8)
    (608,0.7)
    (704,0.6)
    (828,0.8)
    (899,0.8)
    (1108,0.9)
    (1362,0.9)
    (1457,0.85)
    (1559,0.95)
    (3497,0.85)
    (3561,0.7)
    };
    \addlegendentry{LLM 2}
\addplot[
    color=brown!60!black,
    mark=square,
    ]
    coordinates {
    (189,0.95)
    (211,0.9)
    (295,0.9)
    (315,0.8)
    (517,0.9)
    (608,0.8)
    (704,0.7)
    (828,0.9)
    (899,0.9)
    (1108,0.9)
    (1362,0.9)
    (1457,0.9)
    (1488,0.8)
    (1559,0.95)
    (3497,0.9)
    (3561,0.8)
    };
    \addlegendentry{LLM 3}
\addplot[
    color=black,
    mark=square,
    ]
    coordinates {
    (189,0.8)
    (211,0.75)
    (295,0.75)
    (315,0.7)
    (517,0.7)
    (608,0.6)
    (704,0.6)
    (828,0.4)
    (899,0.7)
    (1108,0.85)
    (1362,0.6)
    (1457,0.2)
    (1488,0.7)
    (1559,0.0)
    (3497,0.9)
    (3561,0.6)
    };
    \addlegendentry{LLM 4}
\addplot[
    color=purple,
    mark=square,
    ]
    coordinates {
    (189,0.95)
    (295,0.9)
    (315,0.7)
    (517,0.9)
    (608,0.8)
    (704,0.7)
    (828,0.95)
    (899,0.9)
    (1108,0.9)
    (1362,0.9)
    (1457,0.95)
    (1559,0.95)
    (1939,0.9)
    (3561,0.8)
    };
    \addlegendentry{Federated}
\end{axis}
\end{tikzpicture}
\begin{tikzpicture}
\begin{axis}[
    title={All},
    xlabel={\#words},
    ylabel={Score},
    legend pos=south east,
    ymajorgrids=true,
    grid style=dashed,
    width=.5\textwidth,
    height=.35\textwidth,
    xmode=log
]
\addplot[
    color=blue,
    mark=square,
    ]
    coordinates {
    (189,0.9208)
    (211,0.9333)
    (295,0.801)
    (315,0.8875)
    (517,0.92)
    (608,0.83)
    (704,0.692)
    (828,0.95)
    (899,0.94)
    (1108,0.838)
    (1362,0.9)
    (1457,0.96)
    (1488,0.9)
    (1559,0.933)
    (1592,0.8125)
    (1939,0.9208)
    (3497,0.8917)
    (3561,0.83)
    (18616,0.89375)
    };
    \addlegendentry{LLM 1}
\addplot[
    color=red,
    mark=square,
    ]
    coordinates {
    (189,0.9125)
    (211,0.8458)
    (295,0.648)
    (315,0.8875)
    (517,0.8)
    (608,0.82)
    (704,0.542)
    (828,0.875)
    (899,0.91)
    (1108,0.8333)
    (1362,0.9)
    (1457,0.89)
    (1488,0.88)
    (1559,0.833)
    (1592,0.7875)
    (1939,0.8875)
    (3497,0.7889)
    (3561,0.86)
    (18616,0.86875)
    };
    \addlegendentry{LLM 2}
\addplot[
    color=brown!60!black,
    mark=square,
    ]
    coordinates {
    (189,0.9375)
    (211,0.8792)
    (295,0.79)
    (315,0.9375)
    (517,0.92)
    (608,0.84)
    (704,0.583)
    (828,0.867)
    (899,0.94)
    (1108,0.8167)
    (1362,0.9)
    (1457,0.96)
    (1488,0.858)
    (1559,0.867)
    (1592,0.92)
    (1939,0.9125)
    (3497,0.9167)
    (3561,0.88)
    (18616,0.85)
    };
    \addlegendentry{LLM 3}
\addplot[
    color=black,
    mark=square,
    ]
    coordinates {
    (189,0.9125)
    (211,0.8583)
    (295,0.618)
    (315,0.9125)
    (517,0.82)
    (608,0.87)
    (704,0.575)
    (828,0.842)
    (899,0.75)
    (1108,0.825)
    (1362,0.7)
    (1457,0.81)
    (1488,0.8125)
    (1559,0.65)
    (1592,0.92)
    (1939,0.8625)
    (3497,0.9083)
    (3561,0.89)
    (18616,0.775)
    };
    \addlegendentry{LLM 4}
\addplot[
    color=purple,
    mark=square,
    ]
    coordinates {
    (189,0.9292)
    (295,0.81)
    (315,0.875)
    (517,0.94)
    (608,0.85)
    (704,0.65)
    (828,0.95)
    (899,0.94)
    (1108,0.908)
    (1362,1.0)
    (1457,0.97)
    (1488,0.93)
    (1559,0.933)
    (1592,0.9)
    (1939,0.9417)
    (3561,0.84)
    (18616,0.8875)
    };
    \addlegendentry{Federated}
    \legend{}; 
\end{axis}
\end{tikzpicture}
\caption{Number of words versus scores for single elements of ISAD(G) metadata and overall score}\label{fig:scoresVersusWords}
\end{figure}

\subsection{Analysis and Discussion}\label{sec:analysis}

We have determined the best LLMs for metadata extraction in Sec.~\ref{sec:candidate} and for metadata optimization in Sec.~\ref{federated}.  We will refer to them as LLM 1 (Grok 3 from xAI), LLM 2 (GPT-4-turbo from OpenAI), LLM 3 (DeepSeek-V3 built by DeepSeek), and LLM 4 (Gemini 2.0 Flash from Google) in the analysis. 


We present in Figure~\ref{fig:totaleval} the average LLM scores over all the documents in the dataset for each element, each area, and the whole metadata according to the ISAD(G) standard. Looking at the performance of the single LLMs, LLM 1 performed the best, followed by LLM 3, LLM 2, and finally LLM 4. It is remarkable that the open source LLM 3 is the second-best LLM in our evaluation of the single LLMs. Even the last LLM (LLM 4) also achieves a score of 0.81 for the whole document. Looking at the averages of scores for each element and area, these average scores are never below 0.61. For the best LLM, these average scores are never below 0.78. Finally, the federated approach beats all single LLMs with an average of 0.90 for the whole metadata description, and its average scores for each element and area are never below 0.83.


Table~\ref{tab:Improvements} presents the performance enhancement of our federated approach over directly using single LLMs. For a few elements, such as Id:Extent and AccessUse:Lang, the federated approach does not outperform some individual models. The main reason for this is actually that the federated approach adheres to the archiving standards more strictly and produces more complete content. For example, according to the ISAD(G) specification, the element Id:Extent describes the physical or digital size and format of the unit. For the element, our federated approach creates its content as "3 pages, digital document (PDF)", while the ground truth is only "3 pages". Obviously, the content created by the federated approach is more complete than the ground truth. However, due to the difference between the generated value and the ground truth, the federated approach received a lower score than the LLMs, which extract only "3 pages" for this element, just like the ground truth. Looking at the performance improvements given by the federated approach, coupled with the fact that the federated approach can produce more complete content than the ground truth, we can say that the ability of the federated approach in generation of high-quality metadata is very impressive.



To further analyze the reasons of the superior performance of the federated approach, we also look at the box plots of scores in Figure~\ref{fig:boxplots}, which shows the minimum, maximum, median, 25\%, and 75\% quartiles over all documents. We detect several cases: The federated approach takes over
\begin{itemize}
\item all (or almost all, respectively) the extracted information from the best single LLM (like for the elements Id:Title, Id:DescLev, and AccessUse:Lang (Cont:Scope, respectively)).
\item the best values among the different LLMs (like for the element Id:Date and AccessUse:PhysTech)
\item a mixture of good and bad values (like for the element Id:Extent). In this case the federated approach achieves an average score compared to the scores of the 4 LLMs.
\end{itemize}

Interestingly, we do \emph{not} observe a case where the federated approach takes over the bad values in most cases, and the case, where a mixture of good values and bad values are taken over are rare (i.e., we only observed it for the element Id:Extent), and the reason for this, as discussed earlier, is the capabilities of the federated approach: it can adhere more strictly to the archiving standard and create more exact and complete content than human specialists do.


We present in Figure~\ref{fig:scoresVersusWords} the scores for each element of ISAD(G) metadata versus the number of words of the documents. Interestingly, we see that there are some outliers for documents with around 1,000 words, where the generated metadata descriptions are not similar to the ones of the ground truth. Looking at the overall score for the whole metadata descriptions, we see a tendency for slightly decreasing scores for larger documents (ignoring the outliers with low scores for smaller documents).

\section{Conclusions}
Large Language Models (LLMs) are advanced AI systems that are trained on large amounts of text data and thus acquire knowledge about the world and a remarkable ability to understand natural languages. They are being used to automate a variety of natural language processing tasks such as customer support, content creation, and question answering. This work explores their potential for automated archiving and introduces an LLM-agent-driven AI system for automatic generation of archival metadata.  This system integrates LLM agents with validators, specialized context handling, and a federated optimization technique that unites the intelligence of individual LLMs to produce high-quality metadata descriptions for archival materials.
We conducted an extensive experimental evaluation using real-world archival samples covering documents of various types and data formats. The evaluation results demonstrate the capability of the LLM-based approach in automatic archiving and the superior performance of the federated technique in generating high-quality archival metadata descriptions.


\section*{Acknowledgments}
This work is funded by the German Research Foundation under project number 570892866.

%
%
%
\bibliographystyle{splncs04}
\bibliography{refs}

@inproceedings{groppe2003xpath,
  title={XPath Query Transformation based on XSLT stylesheets},
  author={Groppe, Sven and B{\"o}ttcher, Stefan},
  booktitle={Proceedings of the 5th ACM international workshop on Web information and data management},
  pages={106--110},
  year={2003}
}

@article{groppe2011transforming,
  title={Transforming XSLT stylesheets into XQuery expressions and vice versa},
  author={Groppe, Sven and Groppe, Jinghua and Klein, Niklas and Bettentrupp, Ralf and B{\"o}ttcher, Stefan and Gruenwald, Le},
  journal={Computer Languages, Systems \& Structures},
  volume={37},
  number={2},
  pages={76--111},
  year={2011},
  publisher={Elsevier}
}

@inproceedings{groppe2006satisfiability,
  title={Satisfiability-test, rewriting and refinement of users’ XPath queries according to XML schema definitions},
  author={Groppe, Jinghua and Groppe, Sven},
  booktitle={East European Conference on Advances in Databases and Information Systems},
  pages={22--38},
  year={2006},
  organization={Springer}
}

@article{groppe2006reformulating,
  title={Reformulating XPath queries and XSLT queries on XSLT views},
  author={Groppe, Sven and B{\"o}ttcher, Stefan and Birkenheuer, Georg and H{\"o}ing, Andr{\'e}},
  journal={Data and Knowledge Engineering},
  volume={57},
  number={1},
  pages={64--110},
  year={2006},
  publisher={Elsevier}
}

@inproceedings{groppe2006xpath,
  title={XPath query simplification with regard to the elimination of intersect and except operators},
  author={Groppe, Sven and Bottcher, Stefan and Groppe, Jinghua},
  booktitle={22nd International Conference on Data Engineering Workshops (ICDEW'06)},
  pages={86--86},
  year={2006},
  organization={IEEE}
}

@inproceedings{groppe2005schema,
  title={Schema-based Query Optimization for XQuery Queries.},
  author={Groppe, Sven and B{\"o}ttcher, Stefan},
  booktitle={ADBIS Research Communications},
  year={2005}
}

@inproceedings{groppe2006prototype,
  title={A prototype of a schema-based XPath satisfiability tester},
  author={Groppe, Jinghua and Groppe, Sven},
  booktitle={International Conference on Database and Expert Systems Applications},
  pages={93--103},
  year={2006},
  organization={Springer}
}

@article{groppe2008filtering,
title = {{Filtering Unsatisfiable XPath Queries}},
journal = {Data and Knowledge Engineering},
volume = {64},
number = {1},
pages = {134-169},
year = {2008},
issn = {0169-023X},
__doi = {https://doi.org/10.1016/j.datak.2007.06.018},
url = {https://www.sciencedirect.com/science/article/pii/S0169023X07001395},
author = {Jinghua Groppe and Sven Groppe},
keywords = {Queries, XML, XPath, Satisfiability tester, Query optimization},
abstract = {The satisfiability test checks, whether or not the evaluation of a query returns the empty set for any input document, and can be used in query optimization for avoiding the submission and the computation of unsatisfiable queries. Thus, applying the satisfiability test before executing a query can save processing time and query costs. We focus on the satisfiability problem for queries formulated in the XML query language XPath, and propose a schema-based approach to the satisfiability test of XPath queries, which checks whether or not an XPath query conforms to the constraints in a given schema. If an XPath query does not conform to the constraints given in the schema, the evaluation of the query will return an empty result for any valid XML document. Thus, the XPath query is unsatisfiable. We present a complexity analysis of our approach, which proves that our approach is efficient for typical cases. We present an experimental analysis of our developed prototype, which shows the optimization potential of avoiding the evaluation of unsatisfiable queries.}
}

@book{groppe2008speeding,
  title     = "Speeding up XML Querying",
  author    = "Groppe, Jinghua",
  year      = 2008,
  publisher = "Berlin, Germany: Zugl Lübeck University",
}

@inproceedings{Kessel2025Analysis,
	title = {AI-Supported Analysis and Classification of Digitized Botanical Collections},
	author = {Akasha-Leonie Kessel and Sven Groppe and Dominik Röpert and Jinghua Groppe},
	booktitle = {The 11th International Conference on machine Learning, Optimization and Data science (LOD), Tuscany, Italy},
	year = {2025}
}

@article{Kessel2025Chatbot,
	title = {Impact of Chatbots on User Experience and Data Quality on Citizen Science Platforms},
	author = {Akasha-Leonie Kessel and Soror Sahri and Sven Groppe and Jinghua Groppe and Hanieh Khorashadizadeh and Marc Pignal and Eva Perez Pimparé and Régine Vignes-Lebbe},
	journal = {Computers},
	volume = {14},
	number = {1},
	year = {2025},
	url = {https://doi.org/10.3390/computers14010021}
}

@article{Groppe2020BigData,
	title = {Emergent models, frameworks, and hardware technologies for Big data analytics},
	author = {Sven Groppe},
	journal = {The Journal of Supercomputing},
	volume = {76},
	number = {3},
	pages = {1800-1827},
	year = {2020},
	url = {https://doi.org/10.1007/s11227-018-2277-x}
}

@inproceedings{Junior2024LLMDataPipeline,
	title = {Are Large Language Models the New Interface for Data Pipelines?},
	author = {Sylvio Barbon Junior and Paolo Ceravolo and Sven Groppe and Mustafa Jarrar and Samira Maghool and Florence Sèdes and Soror Sahri and Maurice Van Keulen},
        __author = {Sylvio Barbon Junior and others},
	booktitle = {Proceedings of the International Workshop on Big Data in Emergent Distributed Environments, Santiago, Chile},
	year = {2024},
	__url = {https://doi.org/10.1145/3663741.3664785}
}

@article{ChenLLMEnsembles,
  __author = {Chen,  Zhijun and Li,  Jingzheng and Chen,  Pengpeng and Li,  Zhuoran and Sun,  Kai and Luo,  Yuankai and Mao,  Qianren and Yang,  Dingqi and Sun,  Hailong and Yu,  Philip S.},
  author = {Chen,  Zhijun and others},
  title = {Harnessing Multiple Large Language Models: A Survey on LLM Ensemble},
  journal = {arXiv},
  number = {arXiv:2502.18036},
  year = {2025},
  url = {https://doi.org/10.48550/arXiv.2502.18036}
}

@inproceedings{Ezzabady2024KGConstruct,
	title = {Towards Generating High-Quality Knowledge Graphs by Leveraging Large Language Models},
	author = {Morteza Ezzabady and Frédéric Ieng and Hanieh Khorashadizadeh and Farah Benamara and Sven Groppe and Soror Sahri},
	booktitle = {The 29th Annual International Conference on Natural Language \& Information Systems (NLDB 2024), Turin, Italy},
	year = {2024}
}

@inproceedings{Khorashadizadeh2024Canocanilization,
	title = {Construction and Canonicalization of Economic Knowledge Graphs with LLMs},
	author = {Hanieh Khorashadizadeh and Nandana Mihindukulasooriya and Nilufar Ranji and Morteza Ezzabady and Frédéric Ieng and Jinghua Groppe and Farah Benamara and Sven Groppe},
	booktitle = {International Knowledge Graph and Semantic Web Conference (KGSWC)},
	year = {2024}
}

@misc{Schimmenti24,
  doi = {10.48550/ARXIV.2407.09290},
  author = {Schimmenti,  Andrea and Pasqual,  Valentina and Tomasi,  Francesca and Vitali,  Fabio and van Erp,  Marieke},
  title = {Structuring Authenticity Assessments on Historical Documents using LLMs},
  publisher = {arXiv},
  year = {2024},
}

@misc{Goel23,
  doi = {10.48550/ARXIV.2312.02296},
  __author = {Goel,  Akshay and Gueta,  Almog and Gilon,  Omry and Liu,  Chang and Erell,  Sofia and Nguyen,  Lan Huong and Hao,  Xiaohong and Jaber,  Bolous and Reddy,  Shashir and Kartha,  Rupesh and Steiner,  Jean and Laish,  Itay and Feder,  Amir},
  author = {Goel,  Akshay and others},
  title = {LLMs Accelerate Annotation for Medical Information Extraction},
  publisher = {arXiv},
  year = {2023},
}

@inproceedings{Parekh2023,
  title = {GENEVA: Benchmarking Generalizability for Event Argument Extraction with Hundreds of Event Types and Argument Roles},
  __DOI = {10.18653/v1/2023.acl-long.203},
  booktitle = {Proceedings of the 61st Annual Meeting of the Association for Computational Linguistics},
  publisher = {Association for Computational Linguistics},
  author = {Parekh,  Tanmay and Hsu,  I-Hung and Huang,  Kuan-Hao and Chang,  Kai-Wei and Peng,  Nanyun},
  year = {2023}
}

@inproceedings{Wang2023Code4Struct,
  title = {Code4Struct: Code Generation for Few-Shot Event Structure Prediction},
  __DOI = {10.18653/v1/2023.acl-long.202},
  booktitle = {Proceedings of the 61st Annual Meeting of the Association for Computational Linguistics},
  publisher = {Association for Computational Linguistics},
  author = {Wang,  Xingyao and Li,  Sha and Ji,  Heng},
  year = {2023}
}

@inproceedings{NEURIPS2020_1457c0d6,
 __author = {Brown, Tom and Mann, Benjamin and Ryder, Nick and Subbiah, Melanie and Kaplan, Jared D and Dhariwal, Prafulla and Neelakantan, Arvind and Shyam, Pranav and Sastry, Girish and Askell, Amanda and Agarwal, Sandhini and Herbert-Voss, Ariel and Krueger, Gretchen and Henighan, Tom and Child, Rewon and Ramesh, Aditya and Ziegler, Daniel and Wu, Jeffrey and Winter, Clemens and Hesse, Chris and Chen, Mark and Sigler, Eric and Litwin, Mateusz and Gray, Scott and Chess, Benjamin and Clark, Jack and Berner, Christopher and McCandlish, Sam and Radford, Alec and Sutskever, Ilya and Amodei, Dario},
 author = {Tom Brown and others},
 booktitle = {Advances in Neural Information Processing Systems},
 editor = {H. Larochelle and M. Ranzato and R. Hadsell and M.F. Balcan and H. Lin},
 pages = {1877--1901},
 publisher = {Curran Associates, Inc.},
 title = {Language Models are Few-Shot Learners},
 volume = {33},
 year = {2020}
}

@inproceedings{Khorashadizadeh2024LLMKG,
	title = {Research Trends for the Interplay between Large Language Models and Knowledge Graphs},
	author = {Hanieh Khorashadizadeh and Fatima Zahra Amara and Morteza Ezzabady and Frédéric Ieng and Sanju Tiwari and Nandana Mihindukulasooriya and Jinghua Groppe and Soror Sahri and Farah Benamara and Sven Groppe},
	booktitle = {VLDB 2024 Workshop: The International Workshop on Data Management Opportunities in Unifying Large Language Models + Knowledge Graphs (LLM+KG), Guangzhou, China},
	year = {2024},
	url = {https://vldb.org/workshops/2024/proceedings/LLM+KG/LLM+KG-9.pdf}
}

@book{archives2021using,
 author = "{The National Archives}",
 title = "Using {AI} for Digital Records Selection in Government - Guidance for records managers based on an evaluation of current marketplace solutions ",
 url = "https://cdn.nationalarchives.gov.uk/documents/using-ai-digital-selection-in-government.pdf",
 year = 2021
}

@article{Jaillant2022,
  title = {Applying AI to digital archives: trust,  collaboration and shared professional ethics},
  volume = {38},
  ISSN = {2055-768X},
  __DOI = {10.1093/llc/fqac073},
  number = {2},
  journal = {Digital Scholarship in the Humanities},
  publisher = {Oxford University Press (OUP)},
  author = {Jaillant,  Lise and Rees,  Arran},
  year = {2022},
  month = nov,
  pages = {571–585}
}

@book{Lenartz2022,
  title     = "Digital ist besser? Möglichkeiten der automatisierten Aufbereitung und Bewertung von Fileablagen mit Python am Beispiel einer digitalen Fotosammlung",
  author = {Stephan Lenartz},
  publisher = "Wuerttembergische Landesbibliothek. Stuttgart",
  year      =  2022,
  __url       = "https://doi.org/10.53458/books.105"
}

@article{gillner2023abfragen,
 author = "B. Gillner",
 title = "Abfragen statt anbieten. Eine alternative {Praxis} im archivischen {Umgang} mit {Dateisystemen}.",
 journal = "{Archiv}. {Theorie} und {Praxis}",
 volume = "4",
 pages ={317-–321},
 __url = {https://www.archive.nrw.de/sites/default/files/media/files/Archiv.theoriepraxis23-4-Internet.pdf},
 year = 2023
}

@article{wendt2017,
 author = {G. Wendt and S. Westphal},
 title = {Eine {Herausforderung} des Übergangs: {Fileablagen} als {Quellen} der digitalen Überlieferungsbildung},
 journal = "Transformation ins Digitale. 85. Deutscher Archivtag Karlsruhe",
 pages = {105--113},
 year = 2017
}

@book{naumann2017kreative,
 author = "K. Naumann and M. Puchta",
 title = "Kreative digitale {Ablagen} und die {Archive}",
 publisher = "Ergebnisse eines {Workshops} des {KLA}-{Ausschusses} {Digitale} {Archive} am 22./23 November 2016 in der Generaldirektion der Staatlichen Archive Bayerns. München",
 year = 2017
}

@article{jaillant2022born,
  author = {Lise Jaillant and Katie Aske and Eirini Goudarouli and Natasha Kitcher},
  title = {Introduction: Challenges and Prospects of Born-Digital and Digitized Archives in the Digital Humanities},
  journal = {Archival Science},
  volume = {22},
  pages = {285--291},
  year = {2022},
}

@article{barrueco2022digital,
  author = {José Manuel Barrueco and Miquel Termens},
  title = {Digital preservation in institutional repositories: a systematic literature review},
  journal = {Digital Library Perspectives},
  volume = {38},
  number = {2},
  pages = {161--174},
  year = {2022}
}

\end{document}